\documentclass[leqno]{article}
\usepackage[margin=2cm]{geometry}
\usepackage{fullpage,xcolor,gensymb,graphicx, multicol}
\usepackage{indentfirst, multirow, subcaption}
\usepackage{float}
\usepackage{booktabs}   
\usepackage[title]{appendix}

\usepackage[hebrew,english]{babel}
\usepackage{cjhebrew}

\usepackage{setspace} 

\usepackage{hyperref}
\hypersetup{
    colorlinks=true, 
    linkcolor=blue,  
    urlcolor=blue,   
    citecolor=green  
}

\usepackage{listings}

\definecolor{codegreen}{rgb}{0,0.6,0}
\definecolor{codegray}{rgb}{0.5,0.5,0.5}
\definecolor{codepurple}{rgb}{0.58,0,0.82}
\definecolor{backcolour}{rgb}{0.95,0.95,0.92}
\lstdefinestyle{mystyle}{
    backgroundcolor=\color{backcolour},   
    commentstyle=\color{codegreen},
    keywordstyle=\color{magenta},
    numberstyle=\tiny\color{codegray},
    stringstyle=\color{codepurple},
    basicstyle=\ttfamily\footnotesize,
    breakatwhitespace=false,         
    breaklines=true,                 
    captionpos=b,                    
    keepspaces=true,                 
    numbers=left,                    
    numbersep=5pt,                  
    showspaces=false,                
    showstringspaces=false,
    showtabs=false,                  
    tabsize=2
}
\usepackage{listings}
\usepackage{bera}
\colorlet{punct}{red!60!black}
\definecolor{background}{HTML}{EEEEEE}
\definecolor{delim}{RGB}{20,105,176}
\colorlet{numb}{magenta!60!black}

\title{D-Nikud: Enhancing Hebrew Diacritization with LSTM and Pretrained Models}
\author{Adi Rosenthal and Nadav Shaked\\\texttt{\{adi.rosenthal,nadav.shaked\}@post.idc.ac.il}}
 
\date{August 29th, 2023}

\usepackage{mathtools}

\begin{document}

\maketitle

\begin{multicols}{2}

\section{Abstract}
We introduce D-Nikud, a novel approach to Hebrew diacritization that integrates the strengths of LSTM networks and BERT-based (transformer) pre-trained model. Inspired by the methodologies employed in Nakdimon, we integrate it with the TavBERT pre-trained model, our system incorporates advanced architectural choices and diverse training data. Our experiments showcase state-of-the-art results on several benchmark datasets, with a particular emphasis on modern texts and more specified diacritization like gender.

\section{Introduction}
Diacritization, the process of adding diacritical marks to unvocalized text, is of paramount importance in natural language processing, particularly for languages like Hebrew. In Hebrew, accurate interpretation relies heavily on precise vowel and consonant markings. Beyond enhancing textual clarity, diacritization plays a pivotal role in enabling text-to-vocal reading systems, which are crucial for helping the visually impaired and blind individuals in understanding and accessing written content.

In this paper, we introduce D-Nikud, which seamlessly integrates the strengths of LSTM networks \cite{LSTM} and pre-trained models. Drawing specific inspiration from Nakdimon \cite{Nakdimon} and TavBERT \cite{TavBERT}, our system is designed to address the limitations of existing diacritization models and taps into the contextual knowledge provided by TavBERT, a model pre-trained on a diverse array of textual data. By leveraging this pre-trained model, D-Nikud can identify intricate syntactic and semantic patterns intrinsic to the Hebrew language.

\section{Data}
\subsection{Data Sources}
Our D-Nikud model is trained on data sourced from the Nakdimon \cite{Nakdimon} Training corpora and manually diacritization text by Dicta organization. Given the scarcity of diacritized modern Hebrew text - primarily due to the general practice of reading and writing un-diacritized text - a significant challenge was sourcing appropriate training data. When context fails to provide clarity, only occasional diacritics are added for disambiguation. The data of Nakdimon was compiled from the following sources:

\begin{figure}[H]
\centering
    \begin{tabular}{ |p{3cm}||p{1.3cm}|p{1.7cm}|  }
        \hline
        Genre & Number of Files & Number of Tokens\\
        \hline
        blogs & 46 & 241777\\
        books & 19 & 30537\\
        bsheva & 6 & 22665\\
        critics & 5 & 12785\\
        dont-panic & 75 & 197123\\
        eureka & 10 & 7766\\
        forums & 12 & 21162\\
        govil & 2 & 35000\\
        hadrei-haredim & 4 & 11355\\
        kol & 10 & 21716\\
        kol-briut & 22 & 61601\\
        news & 38 & 158687\\
        nrg & 10 & 4296\\
        president & 10 & 11642\\
        random-internet & 45 & 88135\\
        saloona & 10 & 8781\\
        subs & 10 & 5568\\
        tapuz & 10 & 6225\\
        verdicts & 10 & 7644\\
        wiki & 50 & 301403\\
        yanshuf & 78 & 50542\\
        israel-hayom & 21 & 62862\\
        ynet & 78 & 171715\\
        \hline
        total & 581 & 1540987\\
        \hline
    \end{tabular}
    \captionof{table}{Snapshot of the file and token counts in modern texts by genre. Note that each letter is considered as a token}
\end{figure}

\noindent \textbf{Pre-modern Texts}

\noindent 1. Late pre-modern texts from Project Ben-Yehuda \footnote{\url{https://benyehuda.org}}, primarily composed of works from the late 19th and early 20th centuries.

\noindent 2. Medieval rabbinical texts, with Mishneh Torah \footnote{\url{https://mechon-mamre.org/}} being of notable significance, were obtained from Project Mamre.

\noindent 3. A collection of 23 short stories from the Short Story Project \footnote{\url{https://shortstoryproject.com/he/}}.

\noindent This segment covers roughly 1.81 million Hebrew tokens, with the texts exhibiting varying levels of accuracy, diacritized styles, and degrees of similarity to Modern Hebrew.

\noindent \textbf{Automatically Diacritized Texts}

\noindent We chose not to include this in our training data. This decision was made to preserve the quality and consistency of our dataset.

\noindent \textbf{Modern Texts}

\noindent A manually curated collection of Modern Hebrew texts, which were primarily sourced undiacritized. These texts were diacritized using Dicta \cite{Nakdan} and subsequently manually corrected for errors, either through Dicta's API or via automated scripts designed to identify common mistakes. This dataset, spanning about 1.5M Hebrew tokens, is notably more consistent and aligns closely with the expectations of a native Hebrew speaker compared to the Pre-modern corpus. You can see the modern data details in Table 1.

\noindent \textbf{Dicta Texts}

\noindent In addition to the Nakdimon \cite{Nakdimon} data, we integrated further data supplied by the Dicta organization, including half a million words from standard-score Wikipedia in "ktiv male" format.

\noindent For categorization purposes, we grouped the files into eight main subjects: "dicta", "law", "male\_female", "modern", "poetry", "pre\_modern" and "shortstoryproject\_predotted".

\subsection{Full Script Reconciliation}
When written without the intention of diacritizing, Hebrew often uses a compensatory variant known as full script (Ktiv Male, \<ktyb ml'>). This script introduces instances of the letters '\<y>' and '\<w>' to aid pronunciation, but these can sometimes conflict with standard diacritized script rules. In treating diacritization as a sequence tagging challenge, these additional letters could introduce discrepancies. However, to maintain the integrity of the input, instead of omitting these characters, we adopted a diacritizing policy in line with full script.

\subsection{Character Representations}
In our D-Nikud project, we have created a character representation for the Hebrew letter, partitioning it into three distinct classes: "Nikud", "Dagesh", and "Sin".
The visual representation of our methodology, as can be seen in the image below, offers a clear illustration of how each Hebrew character is dissected and mapped to its corresponding class, optimizing our system's performance in Hebrew diacritization tasks.

\begin{figure}[H]
\centering
    \centering\includegraphics[width=7cm]{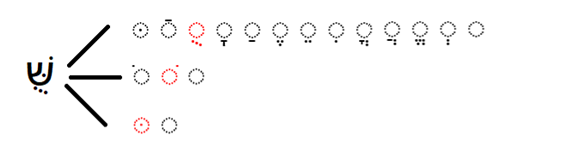}
    \caption{Representation of the D-Nikud Letter class.}
  \label{letter representation}
\end{figure}

The process of Hebrew diacritization is dissected into three distinct classification problems, corresponding to the unique diacritic attributes a Hebrew letter can possess. Each Hebrew character is evaluated for its potential to belong to one or more of these classifications:

\noindent \textbf{"Nikud" Classification:} Addresses the potential vocalization of a character. Letters that can have "Nikud" (vowel diacritics) belong to the set\\
\{ 
'\<t>'-'\<'>', 
'\<k!>', 
'\<n!>' 
\}.

\noindent \textbf{"Dagesh" Classification:} Determines if a letter can have a "Dagesh". This diacritic can be assigned to letters in the set\\
\{ 
'\<b>', 
'\<g>', 
'\<d>', 
'\<h>', 
'\<w>', 
'\<z>', 
'\<.t>', 
'\<y>', 
'\<k\zeronojoin>', 
'\<l>', 
'\<m\zeronojoin>', 
'\<n\zeronojoin>', 
'\<s>', 
'\<p\zeronojoin>', 
'\<.s\zeronojoin>', 
'\<q>', 
'\</s>', 
'\<t>', 
'\<k!>', 
'\<p!>' 
\}.

\noindent \textbf{"Sin" Classification:} Specifically tailored for the Hebrew letter '\</s>', this classification discerns between its two variants, "Sin" ('\<,s>') and "Shin" ('\<+s>'). Only the '\</s>' character is eligible for this classification.

\noindent Note that a letter doesn't necessarily need to have "Nikud", "Dagesh" or "Sin".

\noindent These rules ensure that each Hebrew character is accurately classified, facilitating precise and efficient diacritization.

\subsection{Data processing}
Data processing is a critical component in the pipeline, ensuring the input data is tailored to meet the requirements of the D-Nikud model. We have implemented a suite of classes and methods to streamline this process.

\subsubsection{Read data and combine sentences}
To prepare the data for the D-Nikud model, we follow a specific flow:

\begin{itemize}
    \item Firstly, extracted sentences and their respective labels from the provided file, ensuring only valid Hebrew letters and their diacritics are retained. Any unrelated characters or unneeded diacritics are excluded.
    \item The text is then segmented into sentences, adhering to the model's maximum length constraints. For optimization, we combine these sentences to make the most of the model's input length. This approach is particularly beneficial for shorter sentences, as it minimizes the need for padding. As a result, the model trains and predicts up to six times faster. Additionally, this method aids in better contextual understanding.
\end{itemize}

\subsection{Data Splitting}
For training and evaluation purposes, the data is divided into 90\% training, 5\% validation, and 5\% test sets. By shuffling and distributing the data, we ensure a diverse representation across all sets.

\subsection{Data Tokenization}
We tokenize the input using the TavBert \cite{TavBERT} model's tokenizer. This tokenizer processes each letter, converting it into a one-dimensional representation.

\section{D-Nikud Model}
The D-Nikud model employs a pre-trained TavBERT \cite{TavBERT} Hebrew model for character embeddings, followed by Bi-LSTM layers for sequence processing, a dense layer for further feature refinement, and distinct linear layers for classification outputs.

\noindent \textbf{1. Pre-trained Model:} The model utilizes the pre-trained RobertaForMaskedLM from the TavBERT \cite{TavBERT} Hebrew model as its character embedding layer, devoid of its classification head. Importantly, this embedding layer remains static during the training phase.

\noindent \textbf{2. LSTM Layers:} Subsequent to the TavBERT \cite{TavBERT} model, the architecture incorporates two Bi-LSTM (Bidirectional Long Short-Term Memory) layers. Their bidirectional configuration ensures sequences are processed from both the forward and reverse directions. For model robustness and to curb overfitting, a dropout rate of 0.1 is applied.

\noindent \textbf{3. Dense Layer:} Following the second LSTM's \cite{LSTM} processing, the data is channeled through a dense (fully connected) layer for feature consolidation.

\noindent \textbf{4. Output Layers:} The architecture constructs three specific linear layers. These layers transform features derived from the dense layer into distinct outputs for the "Nikud", "Dagesh" and "Sin" classification tasks.

\begin{figure}[H]
\centering
    \centering\includegraphics[width=7cm]{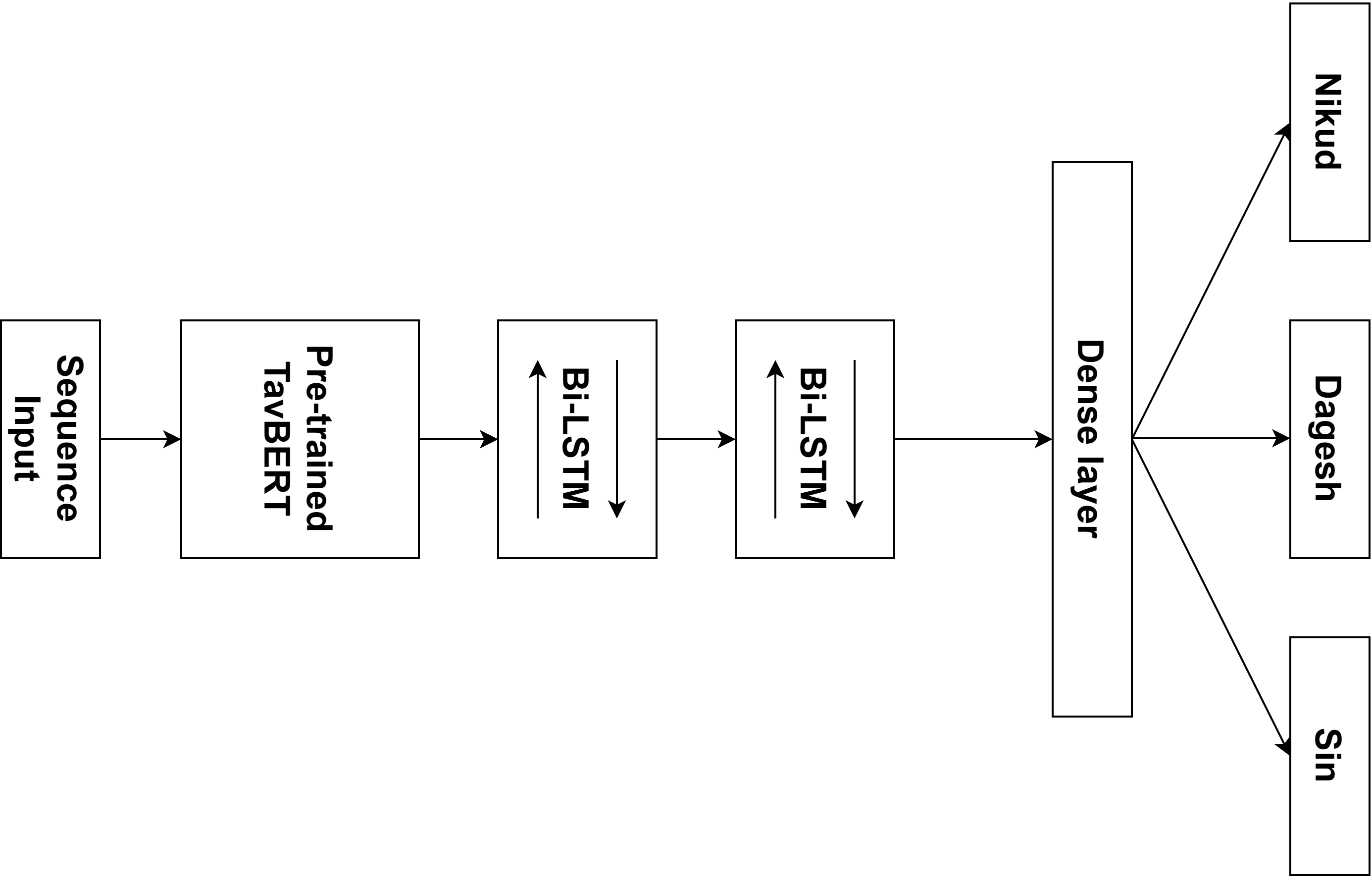}
    \caption{D-Nikud model architecture.}
  \label{architecture}
\end{figure}

\section{Training and Prediction Flow for D-Nikud Model}
\textbf{D-Nikud Model Flow \footnote{\url{https://github.com/NadavShaked/nlp-final-project}}}

\subsection{Training:}
\begin{enumerate}
    \item Acquire a diacritized Hebrew sentence.
    \item Remove diacritics, collecting labels for training.
    \item Combine sentences using padding to match \texttt{MAX\_LENGTH}.
    \item Train the D-Nikud model, adjusting based on prediction errors.
\end{enumerate}

\subsection{Prediction:}
\begin{enumerate}
    \item Input a non-diacritized Hebrew sentence.
    \item The model predicts diacritic labels.
    \item Add predicted diacritic marks to produce a diacritized sentence.
\end{enumerate}

\section{Experiments}
We undertook several experiments to optimize our model and ensure its efficacy:

\noindent \textbf{1. Hyperparameter Tuning:} One of the first challenges we addressed was identifying the optimal set of hyperparameters for our model. Through systematic experimentation, we settled on a learning rate of 0.001, a batch size of 32, and a hidden size of 784. These values were found to deliver the best performance in our tests.

\noindent \textbf{2. Text Combination Analysis:} We investigated the impact of text combination, as detailed in Section 2.3.1, on our model's performance. This experiment aimed to understand the benefits of combining shorter sentences to optimize the model's input length. By comparing the model's performance with and without text combination, we were able to gauge the significance of this preprocessing step.

\noindent \textbf{3. Biblical Text Influence:} Given the distinct linguistic characteristics of biblical Hebrew, we were curious about its influence on a model trained primarily for modern Hebrew diacritization. By training models with and without the inclusion of biblical texts, we discerned that excluding biblical texts led to better performance on modern Hebrew texts. This finding underscores the linguistic evolution of the Hebrew language and its implications for diacritization tasks.

\noindent \textbf{4. Adjusting MAX\_LENGTH Size:} Recognizing the potential impact of input length on model performance, we experimented with varying the MAX\_LENGTH size. Our objective was to determine the ideal sentence length that would both capture the necessary context and ensure efficient processing. Initial results indicated that longer sentences led to increased computational overhead without a corresponding rise in accuracy. After various tests, we decided on a MAX\_LENGTH of 1024 characters. This length balanced computational efficiency with performance, ensuring sentences were not overly elongated, while still retaining the necessary contextual information for accurate diacritization.

\section{Results}

\subsection{Model Convergence Analysis}
Throughout the training phase, we meticulously recorded loss values at each 100 steps intervals, creating a comprehensive training record. Additionally, we conducted an analysis of development accuracy metrics for diacritical marks such as "Nikud," "Dagesh," and "Sin," while also tracking accuracy at the word and letter levels for each epoch. The outcomes are illustrated in the plots below:

\begin{spacing}{0.3}
    \begin{figure}[H]
    \centering
        \centering\includegraphics[width=7cm]{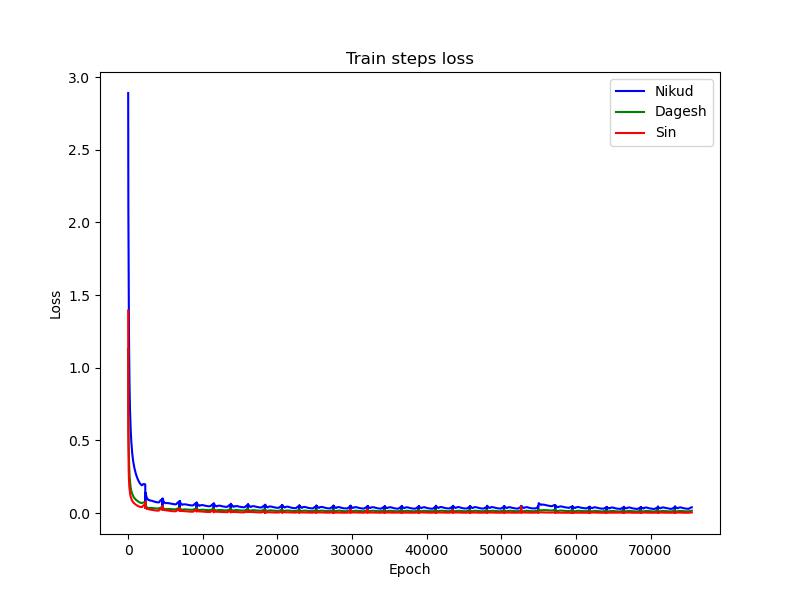}
        \caption{Train set loss by steps of 100 mini-batches plot.}
      \label{Train set loss by steps plot}
    \end{figure}
    
    \begin{figure}[H]
    \centering
        \centering\includegraphics[width=7cm]{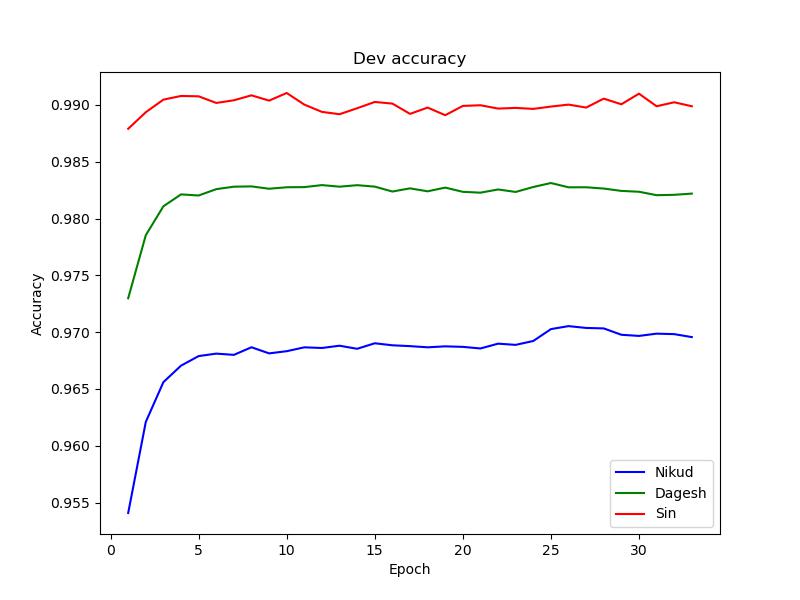}
        \caption{Validation set accuracy by epochs plot.}
      \label{Validation set accuracy plot}
    \end{figure}
    
    \begin{figure}[H]
    \centering
        \centering\includegraphics[width=7cm]{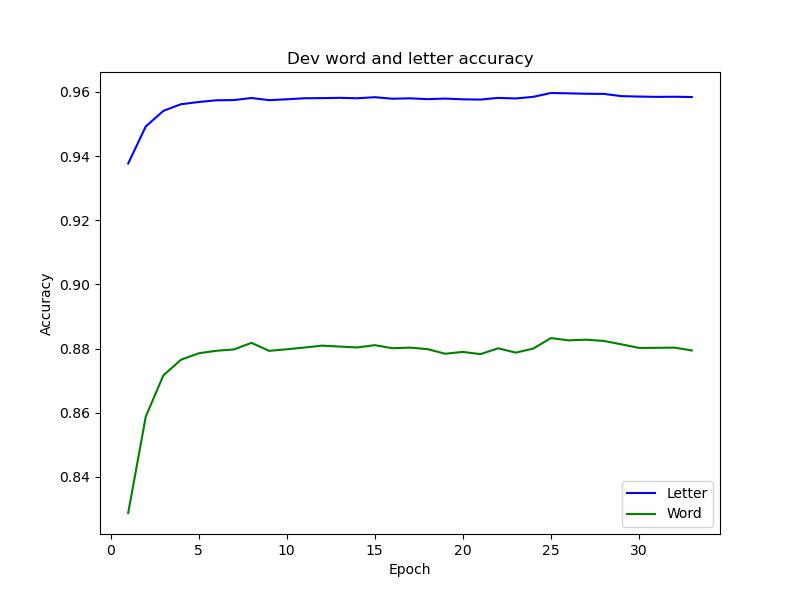}
        \caption{Validation set word and letter accuracy plot.}
      \label{Validation set word and letter accuracy plot}
    \end{figure}
\end{spacing}

\subsection{Test set results}
We extensively tested our model's diacritization accuracy across a range of genres. Our results reveal the model's effectiveness across varied genres.

\begin{figure}[H]
\centering
    \begin{tabular}{ |p{4.5cm}|p{1cm}|p{1cm}|  }
        \hline
        Genre & CHA & WOR \\
        \hline
        dicta & 95.53 & 89.00\\
        law & 98.84 & 96.53 \\
        modern & 97.26 & 91.38\\
        poetry & 95.48 & 88.63 \\
        pre-modern & 94.22 & 83.38\\
        shortstoryproject-predotted  & 97.47 & 92.27 \\
        biblical & 90.69 & 75.74\\
        \hline
    \end{tabular}
    \captionof{table}{D-Nikud test diacritization accuracy across different Hebrew genres (\%).}
\end{figure}

\subsection{Comparative Performance Assessment}
To rigorously assess the efficacy of D-Nikud, we utilized Nakdimon's \cite{Nakdimon} test pipeline. Specifically, we evaluated the performance of D-Nikud on the Nakdimon test set and juxtaposed it with renowned diacritization models: Nakdimon \footnote{\url{https://www.nakdimon.org/}}, Dicta \footnote{\url{https://nakdan.dicta.org.il/}}, Snopi \footnote{\url{https://snopi.com/xPTP/PTP.aspx}}, and Morfix \footnote{\url{https://nakdan.morfix.co.il/}}. The predictions from our model were integrated seamlessly into Nakdimon's established evaluation flow.

\noindent Nakdimon employs a quartet of metrics for evaluation:

\noindent \textbf{1. Decision Accuracy (DEC):} This metric is an aggregate measure, computed over a vast array of individual decisions. It takes into account decisions like dagesh/mappiq for specific letters, sin/shin diacritized for the letter '\</s>', and other pertinent diacritics for eligible letters.

\noindent \textbf{2. Character Accuracy (CHA):} CHA gauges the proportion of characters that are accurately transformed into their intended final form. Notably, a single character's final form might be influenced by multiple decisions, such as a combination of dagesh and vowel.

\noindent \textbf{3. Word Accuracy (WOR):} As the name suggests, this metric reflects the percentage of words in the text that are completely error-free in terms of diacritization.

\noindent \textbf{4. Vocalization Accuracy (VOC):} This unique metric determines the percentage of words where diacritization errors do not lead to mispronunciation among mainstream Israeli Hebrew speakers.

\begin{figure}[H]
\centering
    \begin{tabular}{ |p{1.6cm}||p{1cm}|p{1cm}|p{1cm}|p{1cm}|  }
        \hline
        System & DEC & CHA & WOR & VOC\\
        \hline
        Snopi & 91.29 & 85.84 & 76.45 & 78.91\\
        Morfix & 96.84 & 94.92 & 90.38 & 92.39\\
        Dicta & 97.95 & 96.77 & 94.11 & 94.92\\
        Nakdimon & 97.91 & 96.37 & 89.75 & 91.64\\
        \hline
       \textcolor{red}{D-Nikud} & \textcolor{red}{98.39} & \textcolor{red}{97.15} & \textcolor{red}{90.76} & \textcolor{red}{93.44}\\
        \hline
    \end{tabular}
    \captionof{table}{Document-level macro accuracy for Nakdimon test set (\%).}
\end{figure}

In the table provided (Table 2), various systems are evaluated based on four metrics pertinent to Hebrew diacritization. D-Nikud, highlighted in red, showcases superior performance in both Decision Accuracy (DEC) with a score of 98.26\% and Character Accuracy (CHA) at 96.92\%, surpassing all other systems in these categories. However, for Word Accuracy (WOR) and Vocalization Accuracy (VOC), D-Nikud ranks competitively among the top systems but does not secure the leading position. This comparative analysis emphasizes the strengths of D-Nikud in DEC and CHA while suggesting potential areas for further enhancement in WOR and VOC.

\begin{figure}[H]
\centering
    \begin{tabular}{ |p{1.6cm}||p{1cm}|p{1cm}|p{1cm}|p{1cm}|  }
        \hline
        System & DEC & CHA & WOR & VOC\\
        \hline
        Dicta & 79.56 & 72.01 & 67.33 & 67.75\\
        Nakdimon & 96.92 & 94.61 & 84.87 & 86.05\\
        \hline
       \textcolor{red}{D-Nikud} & \textcolor{red}{97.99} & \textcolor{red}{96.31} & \textcolor{red}{88.45} & \textcolor{red}{91.24}\\
        \hline
    \end{tabular}
    \captionof{table}{Document-level macro accuracy for Female Texts (\%).}
\end{figure}

In Table 3, which compares document-level macro accuracy for texts in the feminine form, D-Nikud stands out by demonstrating superior performance across all metrics when compared to the alternatives. It's worth noting that while the accuracy for feminine language texts is somewhat lower than the general scoring (which predominantly features masculine language), this analysis still underscores D-Nikud's efficacy in diacritization tasks for this category. We posit that our model's capability to account for context plays a crucial role in accurately diacritizing feminine language content. As we gather more data in the future, we anticipate even further improvements in accuracy rates.

\subsection{Computation speed and Efficiency}
A significant advantage of D-Nikud lies in its operational speed. In real-time testing scenarios, D-Nikud consistently demonstrated faster processing times compared to both Dicta \cite{Nakdan} and Nakdimon \cite{Nakdimon}.  This speed advantage positions D-Nikud as a preferable option for applications requiring rapid diacritic predictions without compromising on quality.

\begin{figure}[H]
\centering
    \begin{tabular}{ |p{1.6cm}||p{3cm}|  }
        \hline
        System & Computation time\\
        \hline
        Nakdimon & 2713sec\\
        D-Nikud & 699sec\\
        \hline
    \end{tabular}
    \captionof{table}{Snapshot of the computation time for diacritization of each system}
\end{figure}

\section{Conclusion and Future Work}

Looking ahead, we aim to incorporate an option for diacritization in "Ktiv Haser" (\<ktyb .hsr>). Furthermore, we believe that integrating a word tokenizer alongside the character tokenizer has the potential to produce even more impressive results. By broadening the tokenization approach, we anticipate that D-Nikud's capabilities can be further refined to address intricate diacritization challenges with increased precision and depth.

Additionally, the exploration of refinements in model architecture stands as a promising avenue. Intricate adjustments and the inclusion of more training data could potentially amplify the efficacy of D-Nikud even further.

\section{Related Work}
\subsection{Nakdimon}
\noindent \textbf{Architecture:} Nakdimon \cite{Nakdimon} is built on a two-layer, character-level Long Short-Term Memory (LSTM) network. Despite its architectural simplicity, it performs effectively across a range of Hebrew sources.

\noindent \textbf{Resource Efficiency:} The system proves that accurate Hebrew diacritization can be achieved without extensive human-curated resources, relying solely on diacritized text.

\subsection{Nakdan by Dicta}
\noindent \textbf{Architecture:} Nakdan \cite{Nakdan} combines modern neural models with curated linguistic knowledge, tables, and dictionaries. this combination ensures comprehensive diacritization of Hebrew texts.

\noindent \textbf{Resource Efficiency:} Beyond neural models, Nakdan \cite{Nakdan} emphasizes the importance of incorporating declarative linguistic knowledge. This suggests a balance between automated neural processing and human linguistic expertise.

\subsection{TavBERT}
\noindent \textbf{Character-Based Tokenization:} TavBERT \cite{TavBERT} is a masked language model that employs character tokenization, distinguishing it from models that use subword tokenization.

\noindent \textbf{Training Strategy:} During pretraining, TavBERT \cite{TavBERT} masks random character spans, which the model then attempts to predict, aiming to capture intricate morphological patterns in morphologically-rich languages \cite{MRLs}.

\noindent \textbf{Utilizing:} Utilizing TavBERT's \cite{TavBERT} character-based tokenization, which captures intricate morphological patterns for each character, can potentially enhance the precision and granularity of Hebrew diacritization tasks, offering a comprehensive understanding of the underlying linguistic structures.

\begin{figure}[H]
\centering
    \centering\includegraphics[width=7cm]{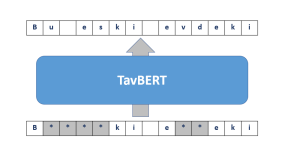}
    \caption{TavBERT model architecture.}
  \label{TavBERT architecture}
\end{figure}

\section{Acknowledgments}
We are sincerely grateful to Elazer Gershony for his invaluable insights and contributions that significantly enriched our project. His expertise and guidance played a crucial role in shaping our work. Additionally, we extend our appreciation to Avi Shmidman for his generous assistance, particularly in providing us with comprehensive Dicta's data \cite{Nakdan}.  We also wish to express our gratitude to the "NLPH - The Natural Language Processing in Hebrew Community" for their assistance in data collection.

\end{multicols}

\clearpage 
\begin{multicols}{2}

\end{multicols}

\end{document}